%% file: neurips_2026.tex
\documentclass{article}

 \usepackage[preprint, nonatbib]{neurips_2026}

\usepackage[utf8]{inputenc} 
\usepackage[T1]{fontenc}    
\usepackage{hyperref}       
\usepackage{url}            
\usepackage{booktabs}       
\usepackage{amsfonts}       
\usepackage{nicefrac}       
\usepackage{microtype}      
\usepackage{xcolor}         
\usepackage{graphicx}
\usepackage{amsmath, amssymb, mathtools, bm, algorithm, algpseudocode}
\usepackage{amsthm, amsmath, amssymb, booktabs, enumitem}

\usepackage{amsthm, amsmath, amssymb}
\newtheorem{theorem}{Theorem}[section]
\newtheorem{proposition}[theorem]{Proposition}
\newtheorem{corollary}[theorem]{Corollary}
\newtheorem{definition}[theorem]{Definition}
\newtheorem{remark}[theorem]{Remark}
\usepackage{booktabs}
\usepackage{multirow}
\usepackage[table]{xcolor}
\usepackage{makecell}

\usepackage{pgfplots}
\usepackage{subcaption}
\usepgfplotslibrary{groupplots}

\title{Elastic Representations via Hyperbolic Geometry}

\author{%
  Arjun Ramesh Kaushik\\
  University at Buffalo, SUNY\\
  \And
  Rudrasis Chakraborty \\
  Lawrence Livermore National Lab \\
  \AND
  Nalini Ratha \\
  University at Buffalo, SUNY\\
  \And
  Venu Govindaraju \\
  University at Buffalo, SUNY \\
}

\pgfplotsset{compat=1.18}
\begin{document}

\maketitle

\begin{abstract}
Learned representations (or embeddings) are the foundation of modern machine learning systems, yet they are typically trained as fixed-size embeddings, without accounting for varying downstream resource or task constraints. Matryoshka Representation Learning (MRL) addresses this limitation by learning nested representations, but requires costly end-to-end retraining and does not support post hoc expansion, while Contrastive Sparse Representations (CSR) reduce training overhead via lightweight adaptors but operate at a 4x larger dimensionality. Additionally, all prior works restrict their framework towards compression only. In this work, we introduce \textbf{H}yperbolic \textbf{E}lastic \textbf{R}epresentation \textbf{L}earning (HERL), a framework for learning dimension-adaptive embeddings (compression and expansion) through a lightweight adaptor without retraining the base encoder. Our key idea is to exploit the exponential capacity of hyperbolic geometry by learning a non-linear mapping from the encoder space to hyperbolic space. We train a simple MLP to downsample the representations before projecting them onto a Lorentzian and learn adaptive representations that respect the Lorentzian geometry. Our approach is model-agnostic and effective in both supervised and unsupervised settings. Through extensive experiments across image and text modalities, we show that HERL preserves downstream performance across a wide range of embedding sizes, exhibits smooth interpolation between dimensions, and achieves these benefits at a fraction of the computational cost. Code: \url{https://github.com/arjunrkaushik/herl}.
\end{abstract}

\section{Introduction}
Machine learning systems encode data into high-dimensional representations that are often assumed to lie on lower-dimensional manifolds. These learned embeddings serve as the foundation for a wide range of downstream tasks. Despite rapid progress in representation learning, most encoders produce fixed-dimensional embeddings, which limits their adaptability in real-world deployments where compute, memory, and latency constraints vary across settings. Matryoshka Representation Learning (MRL) \cite{kusupati2022matryoshka} addresses this limitation by learning nested embeddings that can be truncated at inference time while preserving task performance. However, MRL requires end-to-end retraining of the encoder, which is computationally expensive and often impractical when working with large, pretrained models.

To overcome this, a line of recent work on adaptive representation learning - such as Matryoshka-Adaptor \cite{yoon2024matryoshkaadaptorunsupervisedsupervisedtuning}, SMEC \cite{zhang2025smecrethinkingmatryoshkarepresentation}, and Contrastive Sparse Representations (CSR) \cite{csr} - proposes lightweight adaptors on top of frozen encoders. While these approaches eliminate the need for retraining, they introduce new limitations: SMEC supports only a single reduced dimensionality, CSR activates a subset of dimensions within an overparameterized embedding, and Matryoshka-Adaptor exhibits performance degradation under aggressive compression. In this work, we revisit adaptive representation learning under frozen encoders and formalize two key desiderata: (i) flexible, continuously adjustable embedding dimensionality where the active dimension matches the representation size, and (ii) robustness to aggressive compression with minimal loss in downstream performance. 

The central design problem for an elastic, post-hoc adaptor is that it
must trade dimensionality for fidelity along a continuum: a single
encoder output $z \in \mathbb{R}^D$ must be projected to a representation
$\phi_d(z)$ for any target budget $d \le D$, and the resulting
$d$-dimensional embedding must remain useful downstream. Euclidean
geometry is a poor host for this trade-off, because the volume of a
ball $B_r \subset \mathbb{R}^d$ grows only polynomially as $\Theta(r^d)$, and
embedding a balanced binary tree of depth $\ell$ with bounded distortion already requires dimension
$\Omega(\ell / \log \ell)$~\cite{linial1995geometry,sala2018representation}. As the budget $d$ shrinks, a Euclidean adaptor must therefore
\emph{discard} structure rather than \emph{compress} it. Hyperbolic
geometry inverts this scaling. In the $d$-dimensional hyperbolic space
$\mathbb{H}^d$ of constant negative curvature $c$, the volume of a geodesic ball is
$\mathrm{Vol}(B_r^{\mathbb{H}^d}) \propto e^{(d-1)\sqrt{c}r}$ as $r \to \infty$
\cite{ratcliffe2006foundations}, so capacity grows
\emph{exponentially} in radius~\cite{nickel2017poincare,sala2018representation}. In \cite{sarkar2011low}, the authors
show that \emph{any} finite tree embeds into $\mathbb{H}^2$ with distortion
$1+\epsilon$ for any $\epsilon > 0$, using only $O(\log 1/\epsilon)$
bits of precision that no Euclidean analogue can match at any
dimension. For an adaptor whose budget shrinks toward $d=2$ or $3$, this exponential headroom means that aggressive compression pushes
fine-grained relations outward in radius rather than erasing them.
 
This capacity advantage is paired with a faithfulness advantage on the
structural relations (such as hierarchies, partial orders, graph
distances) that frozen Euclidean encoders typically encode only
implicitly. Bourgain’s theorem shows that embedding an arbitrary $n$-point metric into Euclidean space incurs $O(\log n)$ distortion, with even complete binary trees requiring $\Omega(\sqrt{\log n})$ distortion~\cite{bourgain1985lipschitz,linial1995geometry}. In contrast, the same tree metrics admit near-isometric embeddings in hyperbolic space with distortion $1 + \epsilon$~\cite{sarkar2011low,sala2018representation}. This gap is most relevant for representation learning, where learned structures often resemble hierarchies such as concept trees, taxonomies, and dependency graphs, precisely the regime where Euclidean embeddings are least efficient. Importantly, this is not merely theoretical: pretrained encoders across modalities exhibit low Gromov $\delta$-hyperbolicity~\cite{khrulkov2020hyperbolic}, vision-language models naturally organize representations in hyperbolic-like hierarchies~\cite{desai2023hyperbolic}, and hyperbolic graph networks consistently outperform Euclidean ones on hierarchical benchmarks~\cite{chami2019hyperbolic,liu2019hyperbolic,balazevic2019multi}. Thus, suggesting that the hierarchical structural prior is already baked in the frozen encoder's latent space and the adaptor's role is to \emph{expose} it rather than manufacture it. The hyperbolic embedding additionally endows
coordinates with semantic roles~\cite{nickel2017poincare,tifrea2018poincare,nickel2018learning, chami2021horopca}. Thus, a post-hoc adaptor that respects this
structure can learn nested and elastic embeddings, with minimal degradation in performance at low dimensions.

\begin{figure*}
    \centering
    \includegraphics[width=1\linewidth]{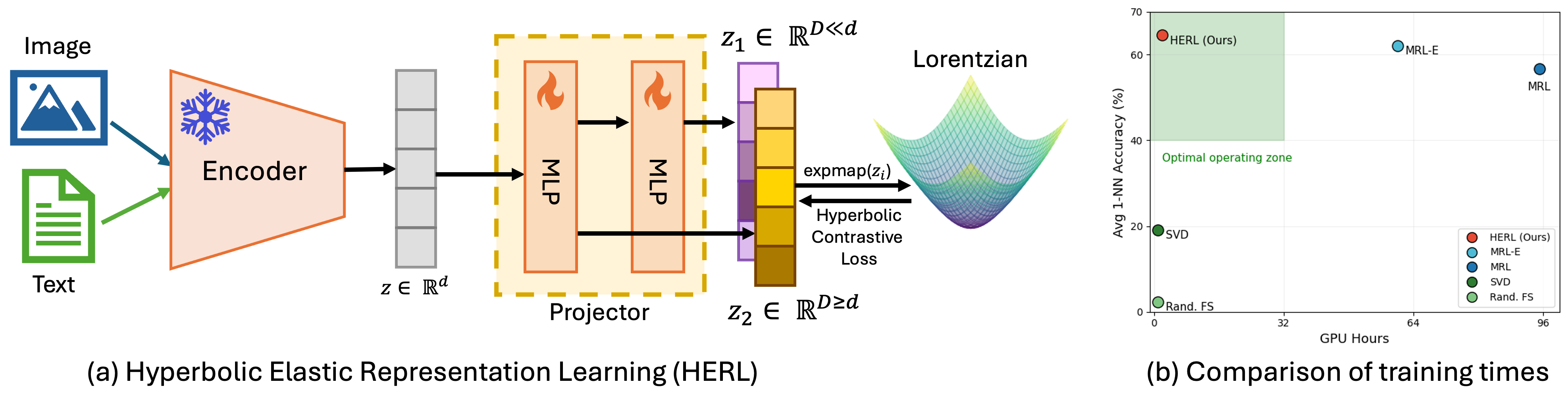}
    \caption{\textbf{Overview.} (a) We propose Hyperbolic Elastic Representation Learning (HERL), a post-hoc adaptor (without encoder retraining) that utilizes the inherent exponential volume capacity and faithful structural preservation properties of hyperbolic geometry. (b) HERL enables elastic representations at a fraction of the training cost. Approximately two hours to train representations of size 8 to 8192, on a single H100 GPU, for the ImageNet-1k dataset. On the other hand, Matryoshka Representation Learning (MRL) needs retraining of the encoder taking about sixty hours on four H100 GPUs.}
    \label{fig:intro}
\end{figure*}

In summary, we propose a novel technique, \textbf{H}yperbolic \textbf{E}lastic \textbf{R}epresentation \textbf{L}earning (HERL), that exploits the exponential volume capacity and structural prior of hyperbolic geometry to learn elastic and nested embeddings without retraining the encoder. HERL satisfies both desiredata, and enables efficient and scalable deployment of pretrained models across diverse resource-constrained environments. Computationally, HERL is approximately 30x cheaper than MRL Fig. \ref{fig:intro}(b).

\section{Background}

\paragraph{Riemannian manifolds.}
A $d$-dimensional Riemannian manifold $(\mathcal{M}, g)$ equips each tangent
space $T_x\mathcal{M}$ with an inner product
$\langle\cdot,\cdot\rangle_x$ that varies smoothly with $x$. The metric induces
the geodesic distance $d_{\mathcal{M}}(x,y)$ as the length of the shortest
curve between $x$ and $y$, together with the
exponential map $\exp_x:T_x\mathcal{M} \to \mathcal{M}$ and its inverse
$\log_x:\mathcal{M} \to T_x\mathcal{M}$, which generalise straight-line
displacement to curved spaces. We refer the reader to \cite{lee2018introduction}
for a comprehensive treatment.

\paragraph{Hyperbolic space.}
\emph{Hyperbolic space} $\mathbb{H}^d$ is the unique simply-connected complete
Riemannian manifold of constant negative sectional curvature
\cite{cannon1997hyperbolic}. Its defining property is that the volume of a
geodesic ball grows \emph{exponentially} with radius rather than polynomially as
in $\mathbb{R}^d$ (Euclidean), thereby giving it sufficient room to host tree-structured data with arbitrarily
low metric distortion: any finite tree embeds isometrically (up to a
$1{+}\varepsilon$ factor) into $\mathbb{H}^2$ \cite{sarkar2011low}. This
property has motivated a substantial line of work using hyperbolic embeddings
to capture latent hierarchies in language, graphs, and vision
\cite{nickel2017poincare,nickel2018learning,ganea2018hyperbolic,
chami2019hyperbolic,khrulkov2020hyperbolic}.
We exploit this property at the level of \emph{representation
compression}: compressing along a hierarchy is far cheaper geometrically when
distances scale exponentially, since coarse and fine semantic levels are
naturally separated by radius rather than competing for orthogonal directions.

\paragraph{The Lorentz model.}
Among the equivalent isometric models of $\mathbb{H}^d$ (Poincaré ball,
Klein, half-space, Lorentz), we work in the Lorentz (hyperboloid) model, which is numerically more stable for gradient-based optimisation than the Poincaré ball \cite{nickel2018learning, law2019lorentzian}.
Let $\langle\cdot,\cdot\rangle_{\mathcal{L}}$ denote the Minkowski bilinear form
on $\mathbb{R}^{d+1}$,
\begin{equation}
\langle x, y \rangle_{\mathcal{L}} \;=\; -x_0 y_0 + \sum_{i=1}^d x_i y_i 
\end{equation}
The Lorentz model with curvature $-c$ ($c>0$) is the upper sheet of the
two-sheeted hyperboloid
$\mathbb{L}^d_c \;=\; \bigl\{\,x \in \mathbb{R}^{d+1} :
   \langle x, x\rangle_{\mathcal{L}} = -1/c,\; x_0 > 0\,\bigr\}$,
endowed with the metric obtained by restricting
$g^{\mathcal{L}} = \mathrm{diag}(-1, 1, \ldots, 1)$ to the tangent space
$T_x \mathbb{L}^d_c = \{v : \langle x, v \rangle_{\mathcal{L}} = 0\}$
\cite{nickel2018learning,law2019lorentzian}. The geodesic distance and
exponential map admit the closed forms
\begin{equation}
\label{eq:lorentz-dist}
d_{\mathbb{L}}(x, y) \;=\;
\frac{1}{\sqrt{c}}\,\mathrm{arcosh}\!\bigl(-c\,\langle x, y\rangle_{\mathcal{L}}\bigr)
\end{equation}
\begin{equation}
\label{eq:lorentz-exp}
\exp^{c}_{x}(v) \;=\;
\cosh\!\bigl(\sqrt{c}\,\|v\|_{\mathcal{L}}\bigr) x
\;+\;
\sinh\!\bigl(\sqrt{c}\,\|v\|_{\mathcal{L}}\bigr)\,
\frac{v}{\sqrt{c}\,\|v\|_{\mathcal{L}}}
\end{equation}
where $\|v\|_{\mathcal{L}} = \sqrt{\langle v, v\rangle_{\mathcal{L}}}$ for any
spacelike $v \in T_x \mathbb{L}^d_c$. Unlike the Poincaré ball
\cite{ganea2018hyperbolic}, the Lorentz model has no conformal factor that
diverges at the boundary, eliminating the dominant source of numerical
instability in hyperbolic deep learning.

\paragraph{Adaptive Representation Learning.} With increasing computational constraints, recent research has pivoted to adaptive representations that can cater to diverse tasks. Early efforts explored context-based architectural adaptations \cite{kim2021lengthadaptivetransformertrainlength}, dynamic widths and depths in BERT \cite{hou2020dynabertdynamicbertadaptive}, and random layer dropping during training to improve pruning robustness \cite{fan2019reducingtransformerdepthdemand}. More recently, Matryoshka Representation Learning (MRL) \cite{kusupati2022matryoshka} introduced a novel technique for creating flexible, nested substructures within embeddings, enabling fine-grained control over the trade-off between latency and accuracy. This concept has since been extended to various modalities and applications, including large language models \cite{nussbaum2025nomicembedtrainingreproducible, yu2024arcticembed20multilingualretrieval}, diffusion models \cite{gu2024matryoshkadiffusionmodels}, and multimodal models \cite{cai2024matryoshkamultimodalmodels, hu2024matryoshkaquerytransformerlarge}. SMEC \cite{zhang2025smecrethinkingmatryoshkarepresentation} and Matryoshka-Adaptor \cite{yoon2024matryoshkaadaptorunsupervisedsupervisedtuning} represent a new trend in Adaptive Representation Learning by introducing a post-hoc adaptor. However, both methods are limited by their generalizability or robustness under aggressive compression. Contrastive Sparse Representations (CSR) \cite{csr} is the first work that offers a post-hoc sparse coding alternative but produces fixed-size representations that are four times bigger than the encoder embedding, limiting its applicability to encoder-dependent downstream tasks. \textit{Towards this end, our method trains a lightweight adaptor on any frozen encoder that explicitly controls the geometric resolution as a function of target dimensionality while remaining adaptable to any dimension.}

\section{Methodology}
\label{sec:method}

We consider a practical post-hoc setting where the encoder \( E: \mathcal{X} \rightarrow \mathbb{R}^D \) is frozen. Given an input \( x \in \mathcal{X} \) with embedding \( z = E(x) \), our goal is to learn a lightweight adaptor \( A: \mathbb{R}^D \rightarrow \mathbb{R}^{D'} \), where $D' \in \mathbb{N}$, such that the transformed representation \( \tilde{z} = A(z) \) yields nested, truncatable embeddings, where any prefix \( \tilde{z}_{1:d} \). To address this challenging problem, we propose a \textbf{H}yperbolic \textbf{E}lastic \textbf{R}epresentation \textbf{L}earning (HERL), which consists of two components: (i) a learnable projector that adapts frozen Euclidean embeddings into a task- and geometry-compatible latent space, and (ii) a hyperbolic optimization module that aligns representations with the Lorentz manifold using contrastive supervision. Together, these components enable the construction of nested embeddings that are both dimensionally truncatable and geometrically coherent, without retraining the base encoder.

\paragraph{(i) Projector ($\mathbf{P}$):}
Since pretrained encoders are generally optimized in Euclidean space, their representations do not naturally align with the geometric properties of hyperbolic manifolds. Therefore, we introduce a learnable projector that serves as a bridge between Euclidean and hyperbolic spaces, enabling a smooth and structured transformation. Concretely, this projector learns to reparameterize the encoder output \( z \in \mathbb{R}^D \) into a form that is compatible with hyperbolic geometry before applying manifold operations.

Beyond geometric alignment, this transformation also addresses a fundamental limitation of high-dimensional embeddings: they often contain redundant or noisy components that are not essential for downstream tasks, especially under strict dimensionality or resource constraints. The projector thus plays a dual role: (i) it filters and compresses the Euclidean representation by selectively preserving task-relevant information, and (ii) it reshapes the representation to better exploit the inductive biases of hyperbolic space, such as its ability to model hierarchical and graph-structured relationships. By jointly performing denoising and geometric adaptation, the projector ensures that the subsequent hyperbolic transformation operates on a compact, informative representation. We realize this projector module P via an MLP as shown in Fig. \ref{fig:intro}(a).

The final layer of P enables elasticity, and is parameterized as \( \mathbb{R}^{D_{\max} \times D} \). For any \( d \leq D_{\max} \), we define: $\bar{z}_{1:d} = (\bar{z}_1, \dots, \bar{z}_d)$, where $\bar{z} = P(E(x)) \cdot W^T$ which corresponds to applying the submatrix \( W_{1:d,:} \). This induces a \emph{prefix-consistent} structure, where lower-dimensional embeddings reuse parameters from higher-dimensional ones. Combined with hyperbolic geometry, this encourages early coordinates to capture global structure (low-radius information) and later coordinates to encode fine-grained details (high-radius refinements), aligning with the exponential volume property.

The key clarification is that our adaptor (Projector and $W$) does not compress in Euclidean space and then separately map into hyperbolic space as two independent stages. The entire adaptor, is trained end-to-end under a loss computed in the Lorentz model, so every gradient it receives flows back through the exponential map and the Lorentzian contrastive objective. The MLP is therefore never optimized for Euclidean reconstruction - it learns to produce tangent-space coordinates whose image under expmap best separates the data on the manifold, meaning compression and geometry are co-designed rather than sequential. This matters because our claim is not that any computation in Euclidean coordinates destroys structure, but the narrower point that an adaptor trained under a Euclidean objective must discard structure as $d$ shrinks, whereas training the same low-dimensional bottleneck against a Lorentzian objective gives the network a target geometry with exponential headroom. Note that the exponential map at the origin is a smooth, invertible reparameterization and thus introduces no information loss.

\paragraph{(ii) Optimization using Hyperbolic Geometry:}
We optimize the adaptor (Projector and $W$) in the Lorentz model by first mapping the Euclidean representation \( \bar{z}_{1:d} \in \mathbb{R}^{d \leq D_{\max}} \) onto the hyperboloid manifold using the exponential map defined at the origin (Eq. \ref{eq:lorentz-exp}):
$u = \text{expmap}(\bar{z}_{1:d}) \in \mathbb{H}^{d \leq D_{\max}}$ where $\mathbb{H}^{D_{\max}} \subset \mathbb{R}^{D_{\max}+1}$. Subsequently, all pairwise similarity computations are performed in hyperbolic space using the Lorentzian metric, ensuring geometric consistency with the manifold structure. Note that the hyperbolic projection is only performed during training. During inference, $\bar{z}_{1:d}$ is directly used for downstream tasks requiring a $d$-dimensional representation.

In the absence of labels, we construct supervision signals from the frozen encoder space. Specifically, for each sample \( x_i \), we utilize its encoder embedding \( z_i = E(x_i) \) and define a neighborhood using cosine similarity (Note the change in notation, $i$ corresponds to sample index and not dimensionality). Subsequetly, we select the top-\(k\) nearest neighbors as positives: $\mathcal{NN}_k(i) = \mathrm{TopK}_{j \neq i}\big(\mathrm{sim}(z_i, z_j)\big)$.
This choice is motivated by the fact that high-dimensional representations from pretrained encoders are approximately concentrated on a hyperspherical manifold, making cosine similarity a more reliable measure of semantic proximity than Euclidean distance.

We define a contrastive objective by treating samples in \( \mathcal{NN}_k(i) \) as positives and all other samples in the batch as negatives. For an anchor \( i \) and a sampled positive \( j \in \mathcal{NN}_k(i) \), the unsupervised InfoNCE-style loss \cite{yue2023hyperboliccontrastivelearning} is given as follows, where \( \tau \) is a temperature parameter:
\[
\mathcal{L}_{\text{unsup}} =
- \log \frac{\exp\left(- d_{\mathbb{H}}(u_i, u_j)/\tau \right)}
{\sum\limits_{k \neq i} \exp\left(- d_{\mathbb{H}}(u_i, u_k)/\tau \right)},
\]
 This formulation leverages the exponential volume growth of hyperbolic space, which naturally supports large separation between dissimilar samples while preserving tight clustering of neighbors, making it particularly suitable for modeling structured semantic relationships. Further, the random sampling over training iterations allows the model to learn faithful local structures.

When class labels \( y_i \) are available, supervision is incorporated through a class-conditioned metric structure in hyperbolic space. Specifically, for an anchor sample \( x_i \), we define its positive set as all samples sharing the same semantic label, $\mathcal{P}(i) = \{ j \neq i \mid y_j = y_i \}$, and implicitly treat all remaining samples in the batch as negatives, i.e., \( \mathcal{N}(i) = \{ k \neq i \mid y_k \neq y_i \} \). This induces a class-aware partitioning of the representation space that aligns with semantic structure. The supervised loss for each anchor \( i \) is then formulated as a normalized log-softmax over the full batch:
\[
\mathcal{L}_{\text{sup}}^{(i)} =
- \frac{1}{|\mathcal{P}(i)|}
\sum_{j \in \mathcal{P}(i)}
\log \frac{\exp\left(-d_{\mathbb{H}}(u_i, u_j)/\tau \right)}
{\sum\limits_{k \neq i} \exp\left(-d_{\mathbb{H}}(u_i, u_k)/\tau \right)}.
\]

Once can interpret this as maximizing the likelihood of correctly identifying class-consistent samples under a hyperbolic Gibbs distribution induced by the Lorentz geometry. The normalization over all non-anchor samples ensures that class separation is globally enforced across the batch, while the averaging over positives prevents collapse in cases of class imbalance or multi-instance classes.

Naively using this opitmization objective across dimensions can cause gradient conflict, leading to a drop in performance across both smaller and larger dimensions. Inspired by SMEC~\cite{zhang2025smecrethinkingmatryoshkarepresentation}, we adopt a sequential training strategy that first optimizes the highest-dimensional representation and progressively refines lower-dimensional embeddings.

\section{Projector Complexity vs.\ Embedding Dimension}
\label{sec:projector-analysis}

We observe empirically that when the target dimension is small ($d \lesssim 32$--$64$),
a lightweight (single-layer) projector P leads to a noticeable degradation in performance,
whereas a deeper multi-layer projector recovers strong results.
Interestingly, beyond this regime, the trend reverses: increasing projector complexity
yields diminishing or even negative returns.
We provide a principled explanation for this phenomenon through three interacting
geometric effects, with detailed proofs and estimators deferred to
Appendix~\ref{supp:projector-analysis}.

\paragraph{Setup.}
Let $E:\mathcal{X}\rightarrow\mathbb{R}^D$ denote a frozen encoder,
and $W \in \mathbb{R}^{D_{\max}\times D}$ the elastic weight matrix.
For an input $x \in \mathcal{X}$, let $z = E(x)$ and define the projected representation $\bar{z}_{1:d} = P(z)\, W_{1:d,:}^{\top}$. Let $\mathcal{S} \subseteq \mathbb{R}^D$ denote the \emph{semantic subspace} induced by $E$,
i.e., the minimal subspace that preserves class-conditional structure, with
$s = \dim(\mathcal{S})$. We identify three failure modes that emerge when $d$ is small:

\paragraph{(i) Rank bottleneck.}
The submatrix $W_{1:d,:}$ has rank at most $d$, which limits the effective degrees of freedom
in the projected space. When $\mathcal{S}$ is not aligned with the coordinate axes and $d < s$,
no linear transformation can fully reorient the representation to align with the dominant
directions of $W$ (see Proposition~\ref{supp:prop:rank}, \ref{supp:sec:rank}).

\paragraph{(ii) Volume collapse.}
The representational capacity of hyperbolic space grows exponentially with dimension:
\[
N(R,\varepsilon,d)\propto
\left(
\frac{\sinh(\sqrt{c}R)}{\sinh(\sqrt{c}\varepsilon)}
\right)^{d-1}
\]
Below a critical volumetric threshold $d_{\mathrm{vol}}$, the space cannot
separate more than $K$ semantic clusters with margin $\varepsilon$ in
$\mathbb{H}^d_c$, causing the geometry to effectively collapse toward
Euclidean behavior (Proposition~\ref{supp:prop:capacity}, \ref{supp:sec:volume}).

\paragraph{(iii) Norm collapse.}
Due to concentration effects in moderate dimensions,
$\mathbb{E}[\|\bar{z}\|_2] \approx \sigma \sqrt{d}$.
When $d < d_{\mathrm{norm}}$, embeddings concentrate near the Lorentz origin,
where hyperbolic distances degenerate:
\[
d_{\mathbb{L}}(u_i,u_j) \approx \|\bar{z}_i - \bar{z}_j\|_2
\]
In this regime, the hyperbolic inductive bias is effectively lost
(Proposition~\ref{supp:prop:norm}, \ref{supp:sec:norm}).

\paragraph{Main result.}
We define the \emph{critical dimension}: $d^* = \max\{s,\, d_{\mathrm{vol}},\, d_{\mathrm{norm}}\}$.

\begin{theorem}[Projector complexity threshold; proof in Appendix~\ref{supp:thm:threshold}]
\label{thm:threshold}
\leavevmode
\begin{enumerate}[label=(\roman*), leftmargin=*, topsep=2pt, itemsep=2pt]
    \item \textbf{Low-dimensional regime ($d < d^*$):} 
    At least one of the failure modes above is active. In this regime,
    a linear projector is insufficient to recover faithful hyperbolic structure,
    while a sufficiently expressive multi-layer projector (depth $\geq 2$, width $\geq D$)
    can jointly mitigate rank misalignment, volume collapse, and norm degeneration.

    \item \textbf{High-dimensional regime ($d \geq d^*$):}
    All three failure modes are inactive. Additional projector capacity
    primarily increases optimization variance without improving expressivity.
    Consequently, simpler projectors are preferred due to reduced overfitting
    and improved stability.
\end{enumerate}
\end{theorem}

\paragraph{Practical estimation.}
For an ImageNet-scale encoder ($D=768$, $K=1000$, $c=1$),
Appendix~\ref{supp:rem:dstar-estimation} provides closed-form estimates of
$s$, $d_{\mathrm{vol}}$, and $d_{\mathrm{norm}}$, yielding
$d^* \approx 32$--$64$, which closely matches our empirical observations
(Section~\ref{sec:experiments}) and aligns with trends reported in
\cite{kusupati2022matryoshka}.

\begin{table*}[t]
\centering
\renewcommand{\arraystretch}{1.25}
\resizebox{\textwidth}{!}{
\begin{tabular}{llcccc|ccc|ccc}
\toprule
\toprule
& & \makecell{Dim} & \multicolumn{3}{c|}{\makecell{\textbf{Text Classification}\\Top-1 Acc (\%)$\uparrow$}}
& \multicolumn{3}{c|}{\makecell{\textbf{Text Clustering}\\Top-1 Acc (\%)$\uparrow$}}
& \multicolumn{3}{c}{\makecell{\textbf{Text Retrieval}\\NDCG@10 (\%)$\uparrow$}} \\
\cmidrule(lr){4-6} \cmidrule(lr){7-9} \cmidrule(lr){10-12}
\textbf{Category} & \textbf{Model} & &
MTOPIntent & Banking77 & TweetSentiment &
BiorxivP2P & BiorxivS2S & TwentyNews &
FiQA2018 & NFCorpus & SciFACT \\
\midrule

Full Rep & NV-Embed-V2 & 4096 
& 93.58 & 92.20 & 79.73
& 53.61 & 49.60 & 64.82
& 62.65 & 43.97 & 77.93 \\
\midrule

\multirow{9}{*}{MRL}
& Stella-1.5B-v5 & 256 
& 90.45 & 86.14 & 76.75
& 50.81 & 46.42 & 60.07
& 55.59 & 36.97 & \textbf{77.48} \\
& Jina-v3 & 256
& 78.81 & 84.08 & 73.81
& 38.14 & 34.39 & 51.96
& 55.73 & 36.63 & 66.63 \\
& Nomic-Embed-V1.5 & 256 
& 72.47 & 83.69 & 59.20
& 38.19 & 31.83 & 48.56
& 35.00 & 32.54 & 68.24 \\
& Gecko-Embed-004(Google) & 256 
& 77.82 & 86.01 & 72.97
& 36.28 & 33.09 & 50.60
& 55.54 & 37.81 & 70.86 \\
& Text-Embed-3-L (OpenAI) & 256 
& 70.45 & 83.19 & 58.98
& 35.43 & 33.86 & 54.24
& 50.33 & 37.94 & 73.10 \\
& Arctic-Embed-L-V2 & 256
& 67.69 & 80.99 & 59.06
& 34.25 & 34.07 & 30.06
& 44.69 & 35.02 & 69.51 \\
& M2V-Base-Glove & 256
& 59.26 & 72.39 & 50.02
& 32.26 & 22.34 & 25.38
& 11.82 & 23.15 & 50.66 \\
& Jina-V3 & 64 
& 68.12 & 67.98 & 71.18
& 36.89 & 33.57 & 50.22
& 44.18 & 33.66 & 68.84 \\
& Nomic-Embed-V1.5 & 64 
& 62.77 & 80.63 & 55.23
& 34.81 & 44.61 & 48.06
& 10.22 & 18.96 & 36.55 \\
& Potion-Base-2M & 64 
& 42.50 & 65.17 & 52.52
& 25.78 & 14.94 & 27.07
& 32.08 & 30.72 & 64.28 \\
\midrule

\rowcolor{gray!15}
Ours & Unsup. HERL & 32 
& 94.39 & 92.88 & 79.51
& 48.18 & 48.47 & 62.85
& 50.70 & 45.67 & 55.70 \\
\rowcolor{gray!15}
(w/ NV-Embed-V2) & Unsup. HERL & 8192 
& 93.72 & 93.21 & 79.34
& 78.43 & 75.22 & 64.70
& \textbf{61.11} & \textbf{55.23} & 71.62 \\
\rowcolor{gray!15}
 & Sup. HERL & 32 
& \textbf{94.53} & \textbf{94.86} & \textbf{79.88}
& 63.28 & 75.82 & 80.14
& - & - & - \\
\rowcolor{gray!15}
 & Sup. HERL & 8192
& 93.49 & 93.15 & 79.60
& \textbf{80.12} & \textbf{76.29} & \textbf{82.77}
& - & - & - \\
\bottomrule
\bottomrule
\end{tabular}}
\caption{\textbf{Results on textual representations.} We compare the downstream utility of representations across three tasks: classification, clustering and retrieval. HERL (Ours) consistently outperforms existing MRL-based representations. Importantly, in some cases, HERL enables a 32 dimensional representation to surpasses full representation (4096 dimension) performance, achieving 128x efficiency in terms of dimension size.}
    \label{tab:text_results}
\end{table*}

\section{Results and Analysis}
\label{sec:experiments}
We evaluate HERL across two modalities - image and text. For image representations, we evaluate the downstream capability of the representations by measuring Top-1 accuracy via linear probing and 1-NN retrieval accuracy on the ImageNet-1k dataset, following MRL \cite{kusupati2022matryoshka}. On the textual representations, we evaluate across three tasks: classification, clustering and retrieval on the MTEB dataset \cite{muennighoff2022mteb}. All implementation and specific dataset details can be found in the appendix. All baseline comparisons directly follow CSR's \cite{csr} protocol. For fair comparisons, we do not include CSR \cite{csr}, whose ambient dimension is consistently 4x the full representation, in our baselines.

\subsection{Image Representations}
\noindent\textbf{Baselines.} We compare our method against several representative approaches: (1) \textbf{MRL / MRL-E} \cite{kusupati2022matryoshka}, which use a ResNet-50 (RN50) backbone where either all nested representations get an independent classifier head (MRL) or a single classifier head is shared amongst nested representations (MRL-E); (2) \textbf{SVD}, where we perform a low-rank approximation of the 1000-class classification layer of RN50 with rank 1000; (3) \textbf{Rand-LP}, which trains a linear classifier on randomly selected feature subsets; and (4) \textbf{Rand-FS}, which randomly subsamples RN50 features for 1-NN classification.

\noindent \textbf{Evaluation metrics.} Following CSR \cite{csr} and MRL \cite{kusupati2022matryoshka}, we adopt 1-NN accuracy, implemented using FAISS \cite{faiss} with exact L2 search. This approach provides an efficient and cost-effective way to evaluate the utility of learned representations for downstream tasks, as 1-NN accuracy requires no additional training. In detail, we use the training set with 1.3M samples as the database and the validation set with 50K samples as the query set. We also report linear probing and few-shot results using Top-1 accuracy.

\begin{figure*}
    \centering
    \includegraphics[width=\linewidth]{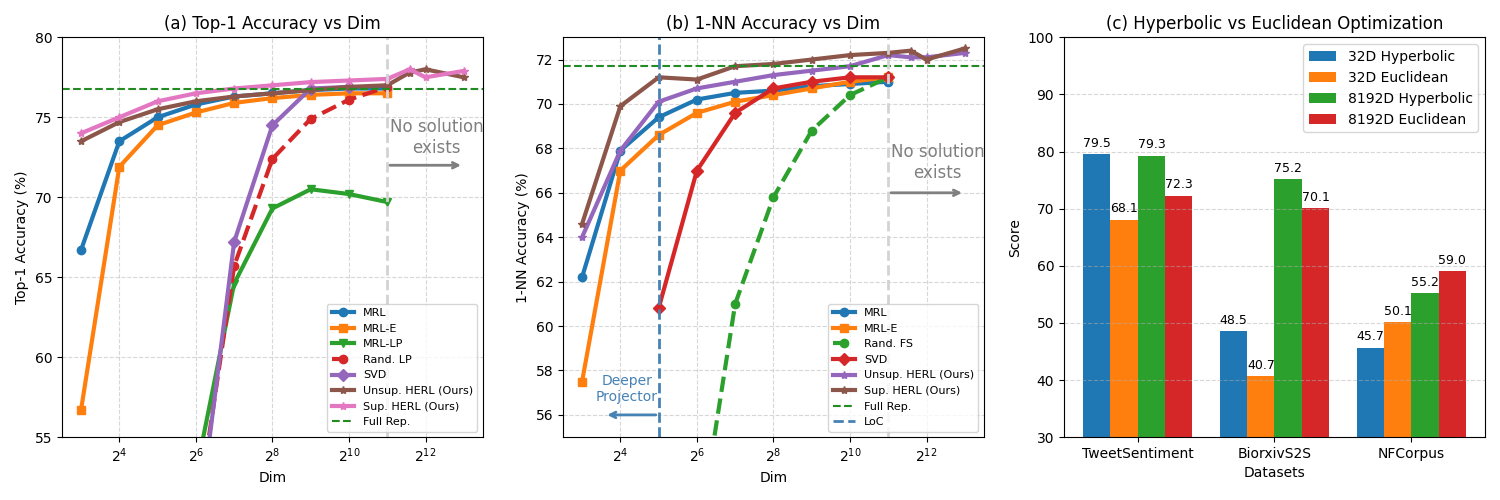}
    \caption{\textbf{Performance of HERL on ImageNet-1k and comparison with Euclidean geometry.} \textbf{(a) Top-1 classification accuracy} and (\textbf{b) 1-NN retrieval accuracy} over embedding dimensions, comparing our supervised and unsupervised HERL adaptors against baselines. The dashed green line ("Full Rep.") denotes the frozen encoder's native-dimension performance. Both HERL variants show improved performance over baselines. The "No solution exists" annotation marks the dimensionality support region that does not exist, whereas HERL extends naturally into this regime. The "Deeper Projector / LoC (Line of Control)" annotation in (b) indicates the dimension below which the adaptor requires a deeper projection head. \textbf{(c) Hyperbolic vs. Euclidean optimization of the adaptor at 32D and 8192D} on three tasks: classification (TweetSentiment), clustering (BiorxivS2S) and retrieval (NFCorpus). Hyperbolic geometry dominates classification and clustering tasks, but struggles with retrieval. Solely due to the fact that the evaluation pipeline normalizes embeddings onto a hypersphere, killing the primal component of distinguishability in hyperbolic geometry - the radius.}
    \label{fig:inet_results}
\end{figure*}

\noindent\textbf{Analysis.} Fig. \ref{fig:inet_results}(a) and Fig. \ref{fig:inet_results}(b) compare HERL against existing baselines across different embedding dimensions for both retrieval Top-1 classification accuracy and 1-NN retrieval accuracy. Traditional projection approaches such as MRL-LP, random linear projections (Rand. LP) exhibit severe degradation under aggressive compression, particularly below 128 dimensions, indicating their inability to preserve semantic structure in compact spaces. While MRL and MRL-E remain competitive at moderate dimensions, they are limited by expensive encoder retraining. In contrast, HERL maintains strong and stable performance across the entire dimensional range, with both unsupervised and supervised variants consistently outperforming competing approaches under extreme compression. Notably, HERL also allows user to go beyond full representation sizes while improving performance. Thus, indicating that our adaptor is learning the latent space rather than merely concatenating zeros. These findings suggest that hyperbolic geometry is an elastic alternative to conventional Euclidean projection methods.

\subsection{Text Representations}
\noindent\textbf{Baselines.} We evaluate several models that provide MRL-style embeddings on the MTEB benchmark \cite{muennighoff2022mteb}, including Stella-1.5B-v5 \cite{zhang2025jasperstelladistillationsota}, Jina-V3 \cite{sturua2024jina}, Nomic-Embed-V1.5 \cite{nussbaum2025nomicembedtrainingreproducible}, GeckoText-Embedding-004 \cite{lee2024geckoversatiletextembeddings}, OpenAI-Text-Embedding-3-L\cite{openai2024embeddings}, Arctic-Embed-L-V2.0 \cite{yu2024arcticembed20multilingualretrieval}, and Potion-Base-2M \cite{minishlab2024model2vec}.

\noindent \textbf{Evaluation Metrics.} We adopt the standard evaluation metrics of the MTEB
benchmark~\cite{muennighoff2022mteb}. Classification and clustering
performance are measured by Top-1 accuracy, while retrieval
performance is measured by nDCG@10~\cite{jarvelin2002cumulated}, the
normalised discounted cumulative gain over the top-10 ranked list,
which is the standard ranking metric in information retrieval and
recommendation.

\noindent \textbf{Analysis.} Table \ref{tab:text_results} shows that HERL achieves strong performance across text classification and clustering tasks while operating at significantly lower dimensions than prior MRL-based approaches. In the highly compressed 32D setting, unsupervised HERL already surpasses / matches the full 4096D NV-Embed-V2 representation on MTOPIntent (94.39 vs.\ 93.58) and TweetSentiment (79.51 vs.\ 79.73), while supervised HERL further improves performance to 94.53 and 79.88 respectively. More importantly, HERL demonstrates exceptional performance on clustering benchmarks, where the supervised 8192D variant achieves state-of-the-art accuracy on BiorxivP2P (80.12), BiorxivS2S (76.29), and TwentyNews (82.77), substantially outperforming both the full representation and all MRL baselines. These results indicate that hyperbolic optimization preserves semantic hierarchy and global structure more effectively than Euclidean methods, particularly under extreme compression. Interestingly, even the compact 32D HERL embeddings remain highly competitive with their 8192D counterparts, suggesting that the hyperbolic adaptor efficiently allocates representational capacity in low-dimensional spaces. HERL exhibits a modest reduction in performance on retrieval tasks (FIQA2018 and SciFACT). We argue that this does not reflect weaker hyperbolic representations, but rather an incompatibility between hyperbolic geometry and conventional retrieval evaluation pipelines. Retrieval protocols projects embeddings onto the unit hypersphere before similarity computation, effectively discarding radial information, which is the principal carrier of semantic separation in hyperbolic space. Thereby, introducing an evaluation bias. This is further elucidated in Sec. \ref{sec:ablation}.

\section{Ablation studies}
\label{sec:ablation}
\paragraph{Why the Lorentz Model?}
 
The Poincar\'e ball, Klein, half-space, and Lorentz models of $\mathcal{H}^d$
are mutually isometric: any geometric quantity computed in one can
in principle be computed in another. Numerically and computationally, however, the
choice matters substantially for an elastic adaptor that must operate
across a wide range of dimensions and embedding magnitudes. We claim
that the Lorentz model is the right choice for our setting, for two concrete reasons.
 
\noindent \underline{(i) Numerical stability away from the boundary:}
Points deep in the hierarchy lie near the boundary of the Poincar\'e
ball, where $\lVert x\rVert_2 \to 1^-$. The hyperbolic distance
involves the factor $1 - \lVert x \rVert^2$ in the denominator, which
suffers catastrophic cancellation in float32 once $\lVert x \rVert > 1 -
10^{-7}$. This is precisely the regime our adaptor must operate in
when $d$ is small and the exponential capacity forces points outward. The Lorentz model
parameterises the same point as $\bm{x} = (x_0, \bm{x}_s) \in \mathbb{R}^{1,d}$
with the constraint $-x_0^2 + \lVert{\bm{x}_s}^2\rVert= -1$ and $x_0 > 0$,
and the distance reduces to
an inner product (Eq. \ref{eq:lorentz-dist}), admitting a stable evaluation up to the limits
of $\mathrm{arcosh}$ itself. Empirically, \cite{nickel2018learning}
demonstrates that Lorentz-model optimization is much more stable than Poincar\'e training, and \cite{mishne2023numerical} formalize this.

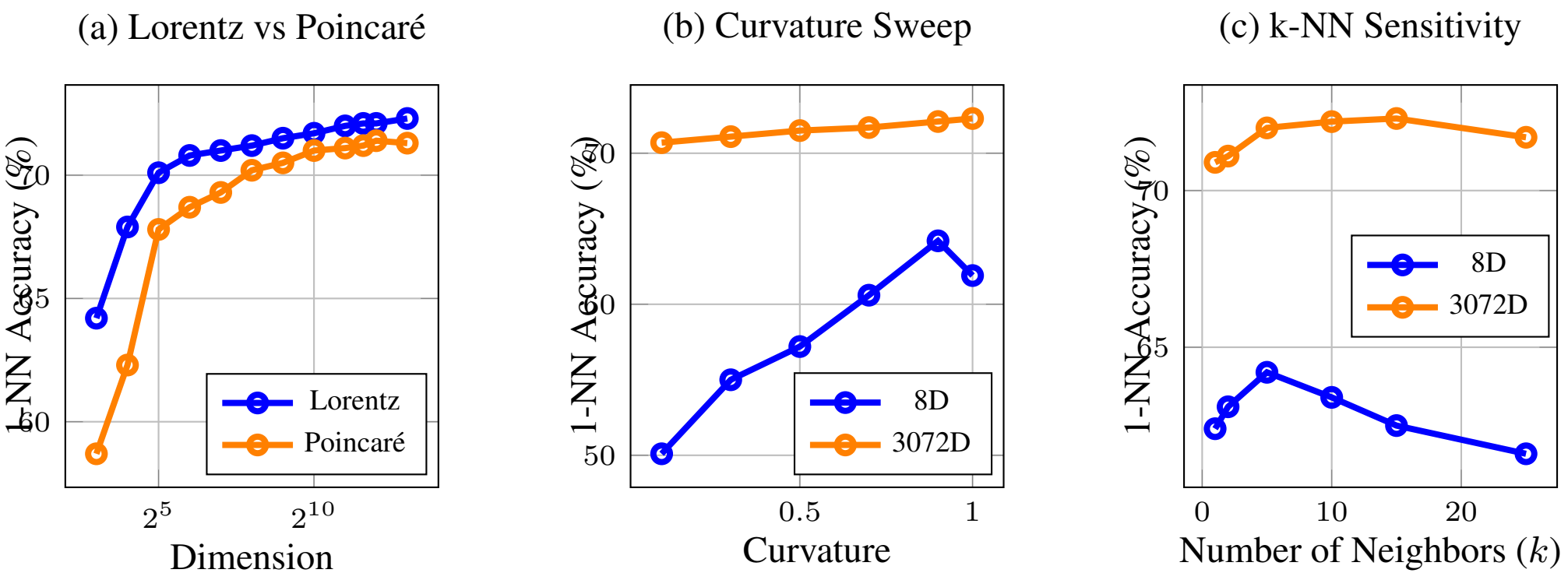
\begin{figure*}[t]
\centering
\begin{tikzpicture}
\pgfplotsset{
    every axis/.append style={
        font=\small,
        label style={font=\small},
        tick label style={font=\scriptsize},
        legend style={font=\scriptsize},
        title style={font=\small},
    }
}

\begin{axis}[
    width=0.34\textwidth,
    height=5cm,
    at={(0,0)},
    xlabel={Dimension},
    ylabel={1-NN Accuracy (\%)},
    xlabel style={yshift=3pt},
    ylabel style={yshift=-12pt},
    title={(a) Lorentz vs Poincaré},
    xmode=log,
    log basis x=2,
    grid=both,
    legend pos=south east
]
\addplot[color=blue, mark=o, line width=1.5pt] coordinates {
(8,64.2) (16,67.9) (32,70.1) (64,70.8)
(128,71.0) (256,71.2) (512,71.5) (1024,71.7)
(2048,72.0) (3072,72.1) (4096,72.1) (8192,72.3)
};
\addlegendentry{Lorentz}

\addplot[color=orange, mark=o, line width=1.5pt] coordinates {
(8,58.7) (16,62.3) (32,67.8) (64,68.7)
(128,69.3) (256,70.2) (512,70.5) (1024,71.0)
(2048,71.1) (3072,71.2) (4096,71.4) (8192,71.3)
};
\addlegendentry{Poincaré}
\end{axis}

\begin{axis}[
    width=0.34\textwidth,
    height=5cm,
    at={(4.8cm,0)},
    xlabel={Curvature},
    ylabel={1-NN Accuracy (\%)},
    xlabel style={yshift=3pt},
    ylabel style={yshift=-12pt},
    title={(b) Curvature Sweep},
    grid=both,
    legend pos=south east
]
\addplot[color=blue, mark=o, line width=1.5pt] coordinates {
(0.1,50.1) (0.3,55.0) (0.5,57.2)
(0.7,60.6) (0.9,64.2) (1.0,61.9)
};
\addlegendentry{8D}

\addplot[color=orange, mark=o, line width=1.5pt] coordinates {
(0.1,70.7) (0.3,71.1) (0.5,71.5)
(0.7,71.7) (0.9,72.1) (1.0,72.3)
};
\addlegendentry{3072D}
\end{axis}

\begin{axis}[
    width=0.34\textwidth,
    height=5cm,
    at={(9.5cm,0)},
    xlabel={Number of Neighbors ($k$)},
    ylabel={1-NN Accuracy (\%)},
    xlabel style={yshift=3pt},
    ylabel style={yshift=-12pt},
    title={(c) k-NN Sensitivity},
    grid=both,
    legend style={at={(0.98,0.5)}, anchor=east}
]
\addplot[color=blue, mark=o, line width=1.5pt] coordinates {
(1,62.4) (2,63.1) (5,64.2)
(10,63.4) (15,62.5) (25,61.6)
};
\addlegendentry{8D}

\addplot[color=orange, mark=o, line width=1.5pt] coordinates {
(1,70.9) (2,71.1) (5,72.0)
(10,72.2) (15,72.3) (25,71.7)
};
\addlegendentry{3072D}
\end{axis}

\end{tikzpicture}
\caption{
\textbf{Ablation studies across geometry, curvature, and neighborhood size.}
(a) Lorentz embeddings outperform Poincaré across dimensions, especially in low-dimensions. (b) Increasing curvature significantly improves compact representations (8D), with limited impact in high dimensions (3072D).
(c) Low-dimensional embeddings are highly sensitive to neighborhood size $k$ that is used in the contrastive loss. All experiments have been performed on ImageNet-1k.
}
\label{fig:ablation}
\end{figure*}

\noindent \underline{(ii) Compatibility with frozen Euclidean encoders:}
The frozen encoder produces $z \in \mathbb{R}^D$ in standard Euclidean
coordinates. A Lorentz-model adaptor lifts $z$ to $\mathbb{H}^d$ via
$\bm{x} = \exp_{\bm{o}}(0, \pi(z))$ where $\bm{o} = (1, \bm{0})$ is
the apex of the hyperboloid and $\pi: \mathbb{R}^D \to \mathbb{R}^d$ is a learned
linear projection; the lift is exact in closed form because the
tangent space at $\bm{o}$ is canonically identified with $\mathbb{R}^d$.
The Poincar\'e lift requires an additional $\tanh$ rescaling that
flattens gradients near the boundary and introduces a second source
of numerical error. Empirically \cite{ganea2018hyperbolic} and
theoretically \cite{shimizu2021hyperbolic}, the Lorentz lift produces
better-conditioned Jacobians. Our experiments  substantiate these insights (Fig. \ref{fig:ablation}(a)).

\noindent \textbf{Euclidean vs. Hyperbolic}
We perform an ablation (Fig. \ref{fig:inet_results}(c)) comparing hyperbolic and Euclidean optimization of the adaptor at both low-dimensional (32D) and full-dimensional (8192D) settings across three representative tasks: classification (TweetSentiment), clustering (BiorxivS2S), and retrieval (NFCorpus). Hyperbolic optimization consistently outperforms Euclidean optimization on classification and clustering tasks, achieving particularly strong gains under extreme compression at 32D (e.g., 79.5 vs. 68.1 on TweetSentiment and 48.5 vs. 40.7 on BiorxivS2S), demonstrating the ability of hyperbolic geometry to preserve semantic and hierarchical structure in compact embedding spaces. Notably, these improvements persist even at 8192D, suggesting that the benefits arise not merely from compression efficiency but from the inductive bias of hyperbolic manifolds themselves, which allocate representational capacity more effectively through exponential volume growth. In contrast, Euclidean optimization slightly outperforms hyperbolic optimization on retrieval (NFCorpus) at both dimensions. We attribute this behavior \textit{not to} inferior hyperbolic representations, but to a mismatch between hyperbolic geometry and standard retrieval evaluation protocols, which normalize embeddings onto the unit hypersphere prior to similarity computation. Such normalization removes radial information, the primary carrier of semantic distinguishability in hyperbolic space, thereby collapsing an important degree of freedom that encodes hierarchy, specificity, and semantic separation. Consequently, existing retrieval pipelines may inherently favor Euclidean embeddings, underestimating the representational quality of hyperbolic embeddings in retrieval settings. That said, retrieval performance in hyperbolic space can be brought closer to the Euclidean by careful tuning of curvature, and temperature in the contrastive loss. The Euclidean experiments have been fine-tuned to extract best performance under the same architectural constraints. 

\noindent \textbf{Effect of curvature}
We ablate the effect of curvature on our unsupervised model in Fig. \ref{fig:ablation}(b) by sweeping values in the range $(0, 1]$. Empirically, higher curvature significantly improves performance in low-dimensional representations, while yielding negligible gains at larger dimensions. This behavior aligns with the exponential capacity of hyperbolic space, which is particularly beneficial in compact embeddings, as well as its ability to capture hierarchical structure. Performance degrades at high curvature ($c=1$) due to a trade-off between hierarchical capacity and local fidelity. While increasing curvature enhances the ability to separate hierarchical structure via exponential volume growth, it also amplifies distances near the boundary, leading to instability and over-separation of nearby points. This degrades local neighborhood preservation, which is critical for nearest-neighbor evaluation. Additionally, extreme curvature introduces optimization challenges and mismatches the partially non-hierarchical structure of real data, resulting in a performance drop. 

\noindent \textbf{Number of Nearest Neighbors}
Fig. \ref{fig:ablation}(c) varies $k$, the number of nearest neighbours from which positives are sampled for the contrastive loss, and exhibits a bias-variance trade-off in the resulting
training signal. At $k{=}1$ each anchor is paired with its nearest point, so the supervisory signal is dominated by whichever neighbour happens to be closest in the frozen encoder's geometry. This magnifies the effect of noise and local distortions, producing high-variance gradients and unstable accuracy. Moderate values $k \in [5, 10]$ aggregate the contrastive signal over several consistent neighbours, dampening the influence of a noisy positive while still concentrating the loss on local structure, and consequently yields peak accuracy. As $k$ grows, the sampled neighbourhood begins to seep across class boundaries. The contrastive objective is then implicitly pulled toward cross-class pairs, biasing the learned geometry toward smoother,
less discriminative embeddings, explaining the small accuracy drop at the right of the curve. The magnitude of this effect exacerbates at $d{=}8$, due to the comparatively limited capacity and accuracy depends strongly on $k$. Higher dimension representations have more room, thereby allowing more robustness across a wide range of $k$. 

\begin{figure}[!t]
\centering
\includegraphics[width=\linewidth]{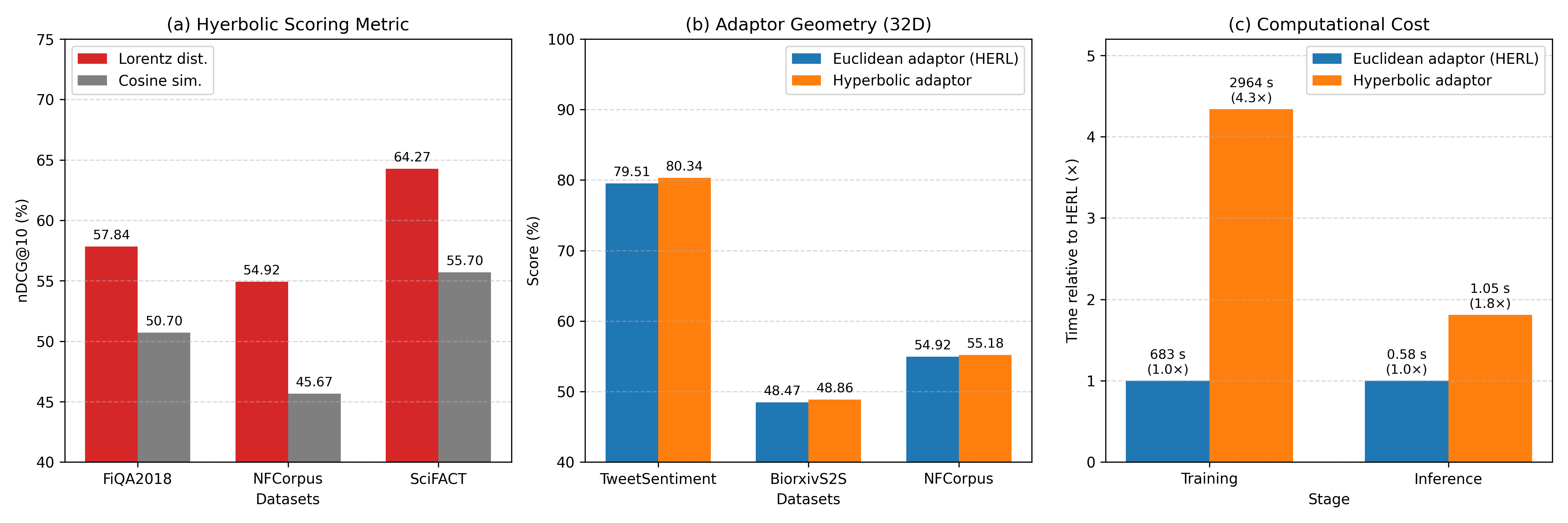}
\caption{\textbf{More ablation studies.}
\textbf{(a) Retrieval with Lorentz distance:} Cosine scoring projects embeddings onto the unit
hypersphere and discards the hyperbolic radius. Evaluation with the native hyperbolic (Lorentz) distance improves results without retraining.
\textbf{(b) Fully hyperbolic adaptor:} As expected a fully hyperbolic adaptor architecture outperforms a Euclidean adaptor, both trained with a hyperbolic objective. However, the negligible gains come at a high computational cost.
\textbf{(c) Computational cost:} The hyperbolic adaptor is $4.3\times$ slower to train and $1.8\times$ slower at inference.}
\label{fig:ablation_bars}
\end{figure}

\paragraph{Retrieval under the native hyperbolic metric.}
Fig. \ref{fig:ablation_bars}(a) revisits the retrieval gap observed in Fig. \ref{fig:inet_results}(c). Standard retrieval protocols rank documents by cosine similarity, which projects every embedding onto the unit hypersphere and discards its norm. For HERL, the norm of determines the hyperbolic radius, which is where the
adaptor places coarse versus fine-grained information. Therefore, we re-evaluate the 32-D unsupervised HERL embeddings, ranking documents by Lorentz distance $d_{\mathcal{L}}$ (Eq. \ref{eq:lorentz-dist}) instead of cosine similarity. No retraining is involved, only the scoring function changes. Scoring with $d_{\mathcal{L}}$ improves HERL across all three retrieval datasets. Thus, the earlier deficit came mainly from the evaluation discarding radial information, not from weaker representations. 

\paragraph{Euclidean versus fully hyperbolic adaptor.}
HERL works with a Euclidean adaptor and uses hyperbolic geometry only in the training loss. A natural alternative is to make the adaptor itself hyperbolic, trained with the same objective. Fig. \ref{fig:ablation_bars}(b,c) compares the two on 32-D embeddings, using
Lorentz distance for retrieval in both cases. 
The fully hyperbolic adaptor gives at most marginal gains at the cost of $4.3\times$ longer training
(2964\,s vs.\ 683\,s) and $1.8\times$ slower inference (1.05\,s vs.\ 0.58\,s per batch of 256, because every layer
requires manifold operations. We conclude that the benefit of hyperbolic geometry comes mainly from the training objective, not from the adaptor's layers.

\section{Conclusion}
In this work, we introduced \textbf{H}yperbolic \textbf{E}lastic \textbf{R}epresentation \textbf{L}earning (HERL), a lightweight and model-agnostic framework for learning dimension-adaptive embeddings without retraining the encoder. Unlike prior approaches that are limited to compression or require expensive end-to-end optimization, HERL supports both compression and expansion through a simple post hoc adaptor trained in an unsupervised manner. By leveraging the exponential representational capacity of hyperbolic geometry and learning adaptive embeddings on the Lorentz manifold, HERL preserves downstream performance across a wide spectrum of embedding dimensions while maintaining smooth interpolation between representations. Most importantly, HERL saves massive hours of encoder retraining. Specifically, one can train a representations of 8 to 8192, in approximately two hours on a single GPU, whereas MRL would take about 60 hours on 4 GPUs. These results highlight the potential of hyperbolic geometry as an effective foundation for scalable and flexible representation learning under varying resource constraints. Extending HERL to additional modalities and more challenging downstream tasks, particularly generative modeling, is an exciting direction for future.

\section*{Acknowledgment}
This work was performed under the auspices of the U.S. Department of Energy by Lawrence Livermore National Laboratory under Contract DE-AC52-07NA27344.

\newpage
\bibliographystyle{splncs04}

\bibliography{reference}

\newpage
\appendix
\section{Full Analysis: Projector Complexity vs.\ Embedding Dimension}
\label{supp:projector-analysis}

This section provides the complete propositions, definitions, corollaries,
theorem, and proof sketch supporting Sec. \ref{sec:projector-analysis} of the main paper.
Throughout, $E:\mathcal{X}\to\mathbb{R}^D$ denotes the frozen encoder, P is the projector MLP, $W \in \mathbb{R}^{D_{\max} \times D}$ the elastic weight matrix,
$z = E(x)$, and $\bar{z}_{1:d} = \text{P}(z)\,W_{1:d}^{\!\top}$ the
$d$-dimensional embedding.

\subsection{Rank Bottleneck and Semantic Subspace Misalignment}
\label{supp:sec:rank}

\begin{proposition}[Rank constraint on linear projections]
\label{supp:prop:rank}
Let $\mathcal{S} \subseteq \mathbb{R}^D$ be the \emph{semantic subspace}
of $E$, i.e.\ the minimal subspace such that the restriction
$E|_{\mathcal{S}}$ preserves class-conditional structure, and let
$s = \dim(\mathcal{S})$.  If $d < s$, then no linear map
$W_{1:d} \in \mathbb{R}^{d \times D}$ can recover $\mathcal{S}$
without information loss.  Moreover, rank sub-multiplicativity
\cite{golub1996matrix} gives
\[
   \mathrm{rank}(W_{1:d} \circ L) \;\leq\;
   \min(d, \mathrm{rank}(L)) \;\leq\; d
\]
for any linear $L \in \mathbb{R}^{D \times D}$, so composing with a
linear P does not increase the effective rank.
When $\mathcal{S}$ is not axis-aligned, the optimal rank-$d$
approximation is the projection onto the leading $d$ eigenvectors of
the data covariance restricted to $\mathcal{S}$ \cite{golub1996matrix}.
\end{proposition}

\begin{corollary}[Necessity of a nonlinear projector at low $d$]
\label{supp:cor:rank}
When $d < s$ and $\mathcal{S}$ is not axis-aligned, a linear P
cannot align $W_{1:d}$ with the semantically relevant directions.
A nonlinear P of depth $\geq 2$ and hidden width $\geq D$ can
approximate the optimal rotation onto $\mathcal{S}$ to arbitrary
precision on compact sets by the universal approximation theorem
\cite{cybenko1989approximation,hornik1991approximation}, reducing the
residual problem for $W$ to a near-axis-aligned projection.
At $d \geq s$, the rank budget exceeds $\dim(\mathcal{S})$ and a linear
P suffices; over-parameterizing P then only introduces optimization
noise.
\end{corollary}

The practical implication is consistent with the theory of Matryoshka
Representation Learning \cite{kusupati2022matryoshka}: coarse prefixes
must encode maximum semantic content, placing the heaviest geometric
burden on $W_{1:d}$ precisely when $d$ is small.

\subsection{Hyperbolic Volume Collapse at Low Dimension}
\label{supp:sec:volume}

The key motivation for working in $\mathbb{L}^d_c$ is the exponential
volume growth of geodesic balls in $\mathbb{H}^d$,
\begin{equation}
\label{supp:eq:vol}
   \mathrm{Vol}_d(r) \;\propto\; \sinh^{d-1}(\sqrt{c}\,r)
   \;\approx\; \tfrac{1}{2^{d-1}}\,e^{(d-1)\sqrt{c}\,r},
   \quad r \to \infty
\end{equation}
compared with $\mathrm{Vol}_d^{\mathbb{R}}(r) \propto r^d$ in Euclidean
space \cite{cannon1997hyperbolic,ratcliffe2006foundations}.
When $d$ is small, the factor $(d-1)$ in the exponent substantially
diminishes this separation advantage.

\begin{proposition}[Effective packing capacity in $\mathbb{H}^d_c$]
\label{supp:prop:capacity}
Fix curvature $c > 0$, ball radius $R > 0$, and separation margin
$\varepsilon > 0$.  The maximum number of $\varepsilon$-separated points
packable in a hyperbolic ball of radius $R$ satisfies
\begin{equation}
\label{supp:eq:packing}
   N(R, \varepsilon, d)
   \;\leq\;
   \left(
      \frac{\sinh(\sqrt{c}\,R)}{\sinh(\sqrt{c}\,\varepsilon)}
   \right)^{d-1}
\end{equation}
following from a standard covering argument using
\eqref{supp:eq:vol} \cite{hamenstaedt1992entropy}.
The ratio $N(R,\varepsilon,d) / N(R,\varepsilon,d')$ grows exponentially
in $d - d'$ for fixed $R, \varepsilon$, confirming that packing capacity
is highly sensitive to $d$ at small values.
\end{proposition}

\begin{definition}[Volumetric dimension threshold]
\label{supp:def:dvol}
Given $K$ classes, the \emph{volumetric threshold} $d_{\mathrm{vol}}$
is the minimum dimension required for $K$ classes to admit an
$\varepsilon$-separated packing inside a hyperbolic ball of radius $R$,
i.e.\ the smallest integer $d$ satisfying $N(R,\varepsilon,d) \geq K$.
Applying \eqref{supp:eq:packing} and solving yields
\begin{equation}
\label{supp:eq:dvol}
   d_{\mathrm{vol}}
   \;=\;
   1 \;+\;
   \left\lceil
      \frac{\log K}
           {\log\sinh(\sqrt{c}\,R) - \log\sinh(\sqrt{c}\,\varepsilon)}
   \right\rceil
\end{equation}
\end{definition}

\begin{corollary}[Diminishing geometric margin at low $d$]
\label{supp:cor:volume}
For $d < d_{\mathrm{vol}}$, $\mathbb{H}^d_c$ cannot geometrically
accommodate $K$ $\varepsilon$-separated class clusters regardless of
$P$.  The loss landscape becomes geometrically closer to Euclidean, and
the exponential inter-class separation that motivates hyperbolic
embeddings is substantially weakened.  A complex projector compensates
by engineering explicit class separation \emph{before} the exponential
map, so that the limited hyperbolic volume is used efficiently.
\end{corollary}

This is consistent with \cite{sala2018representation}, who demonstrate
that hyperbolic embedding fidelity degrades rapidly below $d\approx
10$--$50$ for tree-structured data, and with \cite{gu2019learning}, who
observe that effective curvature must be increased to compensate for
small $d$.

\subsection{Norm Collapse and the Flat Neighbourhood of the Origin}
\label{supp:sec:norm}

The Lorentz exponential map at the origin
$o = (1/\sqrt{c},\,\mathbf{0}) \in \mathbb{L}^d_c$ is
\begin{equation}
\label{supp:eq:expmap-origin}
   \exp^c_o(v)
   \;=\;
   \cosh\!\bigl(\sqrt{c}\,\|v\|_\mathcal{L}\bigr)\,o
   \;+\;
   \sinh\!\bigl(\sqrt{c}\,\|v\|_\mathcal{L}\bigr)\,
   \frac{v}{\sqrt{c}\,\|v\|_\mathcal{L}}
\end{equation}
where $\|v\|_\mathcal{L} = \sqrt{\langle v,v\rangle_\mathcal{L}}$ controls
the radial position on $\mathbb{L}^d_c$.

\begin{proposition}[Norm concentration at low dimension]
\label{supp:prop:norm}
Let $\bar{z} \in \mathbb{R}^d$ have i.i.d.\ coordinates with zero mean
and variance $\sigma^2$.  By concentration of measure
\cite{vershynin2018highdimensional},$
   \bigl|\,\|\bar{z}\|_2 - \sigma\sqrt{d}\,\bigr|
   \;\leq\; O(\sigma)$ with high probability,
so $\mathbb{E}[\|\bar{z}\|_\mathcal{L}] \approx \sigma\sqrt{d}$.
For $d \ll D$, this quantity is small, placing
$\exp^c_o(\bar{z})$ near $o$ where the sectional curvature experienced
by the embedding is negligible \cite{law2019lorentzian}.
\end{proposition}

\begin{table*}[t]
\centering
\renewcommand{\arraystretch}{1.3}
\resizebox{\textwidth}{!}{%
\begin{tabular}{llll}
\toprule
\toprule
\textbf{Failure mode}
  & \textbf{Active when}
  & \textbf{Root cause}
  & \textbf{Fix (low $d$)} \\
\midrule
Rank bottleneck
  & $d < s$
  & Linear $P$ cannot rotate $\mathcal{S}$ to axis-aligned
  & Deep nonlinear $P$ \\
Volume collapse
  & $d < d_{\mathrm{vol}}$
  & $K$ classes not $\varepsilon$-separable in $\mathbb{H}^d_c$
  & Class-aware $P$ \\
Norm collapse
  & $d < d_{\mathrm{norm}}$
  & $\|\bar{z}\|_2 < r^*$; geometry degenerates to Euclidean
  & Scale-amplifying $P$ \\
\bottomrule
\bottomrule
\end{tabular}}
\caption{Summary of the three failure modes, their dependence on target
dimension $d$, and the corresponding projector remedy.}
\label{supp:tab:failure-modes}
\end{table*}

\begin{definition}[Norm dimension threshold]
\label{supp:def:dnorm}
Let $\delta > 0$ control the minimum acceptable non-Euclidean
distortion.  Define the \emph{critical radius}
\begin{equation}
\label{supp:eq:rstar}
   r^* \;=\; \frac{1}{\sqrt{c}}\,\mathrm{arccosh}(1 + \delta)
\end{equation}
the boundary of the flat neighbourhood of $o$.  The
\emph{norm threshold} is
\begin{equation}
\label{supp:eq:dnorm}
   d_{\mathrm{norm}}
   \;=\;
   \left\lceil \frac{(r^*)^2}{\sigma^2} \right\rceil
   \;=\;
   \left\lceil
      \frac{\mathrm{arccosh}^2(1+\delta)}{c\,\sigma^2}
   \right\rceil
\end{equation}
For $d < d_{\mathrm{norm}}$, we have
$\mathbb{E}[\|\bar{z}\|_2] < r^*$, so embeddings are expected to lie
inside the flat neighbourhood.
\end{definition}

\begin{remark}[Degeneration to Euclidean metric in the flat neighbourhood]
\label{supp:rem:flat}
For $\|\bar{z}\|_\mathcal{L} < r^*$, the Taylor expansions
$\cosh(r) \approx 1 + r^2/2$ and $\sinh(r) \approx r$ give
$\exp^c_o(\bar{z}) \approx o + \bar{z}$, so the Lorentz geodesic
distance reduces to $d_{\mathbb{L}}(u_i, u_j) \approx
\|\bar{z}_i - \bar{z}_j\|_2$.
The contrastive loss therefore carries no hyperbolic inductive bias in
this regime, negating the motivation for the Lorentz model.
\end{remark}

\begin{corollary}[Projector must amplify norms at low $d$]
\label{supp:cor:norm}
For $d < d_{\mathrm{norm}}$, a complex $P$ must explicitly scale
embeddings so that $\|P(z)\|_\mathcal{L} \geq r^*$.  A linear layer
will not learn this non-trivial amplification reliably, because the
contrastive loss gradient does not distinguish Euclidean-regime collapse
from correct hyperbolic placement when embeddings are near $o$.
At $d \geq d_{\mathrm{norm}}$, the norm $\sigma\sqrt{d} \geq r^*$
intrinsically, so a simple projector suffices.
A practical diagnostic is to monitor $u_0$, if $u_0 \approx 1/\sqrt{c}$ at low $d$, norm collapse is confirmed.
\begin{equation}
\label{supp:eq:x0-diagnostic}
   u_0 \;=\; \cosh\!\bigl(\sqrt{c}\,\|\bar{z}\|_\mathcal{L}\bigr)
   \;\gg\; \frac{1}{\sqrt{c}};
\end{equation}
\end{corollary}

\subsection{Formal Threshold Theorem and Proof}
\label{supp:sec:threshold}

\begin{theorem}[Projector complexity threshold]
\label{supp:thm:threshold}
Let $s = \dim(\mathcal{S})$ with $\mathcal{S}$ not axis-aligned,
and let $d_{\mathrm{vol}}$, $d_{\mathrm{norm}}$ be as in
Definitions~\ref{supp:def:dvol}--\ref{supp:def:dnorm}.
Define the \emph{critical dimension}
\begin{equation}
\label{supp:eq:dstar}
   d^* \;=\; \max\bigl(s,\; d_{\mathrm{vol}},\; d_{\mathrm{norm}}\bigr)
\end{equation}
\begin{enumerate}[label=(\roman*)]
  \item \textbf{Necessity of a complex projector below $d^*$.}
        For any $d < d^*$, a linear P is insufficient: at least one
        of the following active failure modes prevents faithful
        hyperbolic embedding regardless of $W$:
        \begin{itemize}
          \item \emph{Rank bottleneck:}
                $\mathrm{rank}(W_{1:d} \circ L) \leq d < s$ for any
                linear $L$ (Corollary~\ref{supp:cor:rank});
          \item \emph{Volume collapse:}
                $N(R,\varepsilon,d) < K$
                (Corollary~\ref{supp:cor:volume});
          \item \emph{Norm collapse:}
                $\mathbb{E}[\|\bar{z}\|_2] < r^*$, so
                $d_{\mathbb{L}} \approx \|\cdot\|_2$
                (Corollary~\ref{supp:cor:norm}).
        \end{itemize}
        A projector of depth $\geq 2$ and hidden width $\geq D$
        resolves all three by the universal approximation theorem
        \cite{cybenko1989approximation,hornik1991approximation}.

  \item \textbf{Sufficiency of a simple projector above $d^*$.}
        For $d \geq d^*$, all failure conditions are absent.
        The SGD bias-variance decomposition \cite{bottou2012stochastic}
        gives $\mathrm{MSE}(\hat{g}) = \mathrm{bias}^2 +
        \sigma^2_{n_P}/N$ where $\mathrm{bias}=0$ and
        $\sigma^2_{n_P}$ grows monotonically with projector parameter
        count $n_P$; minimising $n_P$ therefore minimises total error.
\end{enumerate}
\end{theorem}

\begin{proof}
\textit{(i) Necessity.}
The three failure modes are independent; $d < d^*$ activates at least
one by definition of $d^*$ as their minimum.
For the rank bottleneck: if $d < s$, rank sub-multiplicativity
\cite{golub1996matrix} gives
$\mathrm{rank}(W_{1:d} \circ L) \leq d < s$ for any linear $L$, so a
linear P cannot align $W_{1:d}$ with $\mathcal{S}$.
For volume collapse: if $d < d_{\mathrm{vol}}$,
Definition~\ref{supp:def:dvol} yields $N(R,\varepsilon,d) < K$ directly
from \eqref{supp:eq:packing}, so $K$-class separation does not exist in
$\mathbb{H}^d_c$ regardless of P.
For norm collapse: if $d < d_{\mathrm{norm}}$,
Definition~\ref{supp:def:dnorm} gives
$\mathbb{E}[\|\bar{z}\|_2] < r^*$, and Remark~\ref{supp:rem:flat} shows
the Lorentz metric degenerates to Euclidean.
A depth $\geq 2$, width $\geq D$ projector resolves each failure mode by
the universal approximation theorem \cite{hornik1991approximation}.

\textit{(ii) Sufficiency.}
When $d \geq d^*$, all three failure conditions are absent by definition.
The bias is therefore zero and the $\mathcal{O}(n_P/N)$ variance term
from \cite{bottou2012stochastic} grows with projector capacity, making a
simple $P$ strictly preferable.
\end{proof}

\begin{remark}[Practical estimation of $d^*$]
\label{supp:rem:dstar-estimation}
Given a pretrained encoder $E$, $s$ can be estimated as the numerical
rank of the empirical covariance
$\hat{\Sigma} = \frac{1}{N}\sum_i z_i z_i^\top$ at a relative tolerance
of $10^{-2}$ \cite{golub1996matrix}.
$d_{\mathrm{vol}}$ is computable from $K$, $R$ (set to the $95$th
percentile of $\|u_i\|$ in a pilot run), and $\varepsilon$ (set to the
minimum intra-class geodesic distance in a pilot run).
$d_{\mathrm{norm}}$ requires only $c$, $\sigma^2$ (estimated from pilot
embeddings), and tolerance $\delta$ (we suggest $\delta=0.1$).
For ImageNet-scale encoders with $D=768$, $K=1000$, and $c=1$, these
formulas yield $d^*\approx 32$--$64$, consistent with the experimental
observations in Sec. \ref{sec:experiments} of the main paper.
\end{remark}

\begin{table*}[!b]
    \centering
    \renewcommand{\arraystretch}{1.3}
    \resizebox{\textwidth}{!}{
    \begin{tabular}{c|ccccccc|c}
    \toprule
    \toprule
         \textbf{Dim} & \textbf{Full Rep.} & \textbf{Rand. FS} & \textbf{SVD} & \textbf{MRL} & \textbf{MRL-E} & \textbf{Unsup. HERL} & \textbf{Sup. HERL} & \textbf{CSR} \\
         \midrule
        8 & - & 2.36 & 19.14 & 62.19 & 57.45 & 63.90 & \textbf{64.60} & 67.78\\
        16 & - & 12.06 & 46.02 & 67.91 & 67.05 & 67.89 & \textbf{69.90} & 69.17 \\
        32 & - & 32.91 & 60.78 & 69.46 & 68.60 & 70.20 & \textbf{71.23} & 70.15\\
        64 & - & 49.91 & 67.04 & 70.17 & 69.61 & 70.71 & \textbf{71.14} & 70.94\\
        128 & - & 60.91 & 69.63 & 70.52 & 70.12 & 71.00 & \textbf{71.72 }& 70.99\\
        256 & - & 65.75 & 70.67 & 70.62 & 70.36 & 71.20 & \textbf{71.89 }& 71.31\\
        512 & - & 68.77 & 71.06 & 70.82 & 70.74 & 71.52 & \textbf{72.05} & 71.29\\
        1024 & - & 70.41 & 71.22 & 70.89 & 71.07 & 71.70 & \textbf{72.20} & 71.30\\
        2048 & 71.19 & 71.19 & 71.21 & 70.97 & 71.21 & 72.00 & \textbf{72.29} & 71.20\\
        3072 & - & - & - & - & - & 72.11 & \textbf{72.49} & 72.04\\
        4096 & - & - & - & - & - & \textbf{72.19} & 72.09 & 71.87\\
        8192 & - & - & - & - & - & 72.34 & \textbf{72.56} & 72.11\\
        \bottomrule
        \bottomrule
    \end{tabular}}
    \caption{\textbf{1-NN accuracy.} Comparison of different methods on ImageNet-1k for 1-NN retrieval accuracy. Note that, CSR \cite{csr} uses an ambient dimension of 8192 (4x), activating only certain dimensions. Hence, has been omitted from comparison. The highest score has been \textbf{bolded}.}
    \label{tab:appndx_image_1nn}
\end{table*}
\section{Experiments on image representations}

\subsection{Datasets} 

We utilize ImageNet-1K \cite{imagenet-1k}, a large-scale visual dataset widely used for developing and benchmarking computer vision models. It comprises 1,000 object categories, each containing a diverse collection of images. In total, the dataset includes 1,281,167 training images, 50,000 validation images, and 100,000 test images.

\subsection{Implementation details}
We follow MRL \cite{kusupati2022matryoshka}, and utilize ResNet50 weights to extract encoder representations for our HERL model, adopting the same preprocessing procedures as well. The projector module is realized through a small MLP consisting of four layers ($2048 \to 1024 \to 512 \to 256$). The output of the projector is used to learn elastic representations. All experiments have been performed on a single H100 GPU. Note that, we also clip the norm of the representations to $5.0$ before Lorentzian projection to avoid numerical instability errors. For larger dimensions, we perform graident accumulation to allow operating at the batch size. All hyperbolic operations have been performed using the $\mathrm{geoopt}$ library. See Table \ref{tab:image_impl_details} for more specific details.

\begin{table*}[!b]
    \centering
    \renewcommand{\arraystretch}{1.3}
    \resizebox{\textwidth}{!}{
    \begin{tabular}{ccccccc}
    \toprule
    \toprule
    Backbone & Input dim & LR & Epochs & Batch size & Optimizer & Curvature \\
    \midrule
   ResNet50 & 2048 & 1e-3 & 10 & 2048 & RiemannianAdam & 0.9 \\
    \bottomrule
    \bottomrule
    \end{tabular}}
    \caption{Implementation details for experiments on image representations with ImageNet-1k}
    \label{tab:image_impl_details}
\end{table*}

\subsection{Results}
Exact numerical values for Top-1 classification accuracy and 1-NN retrieval are shown in Table  \ref{tab:appndx_image_cls} and Table \ref{tab:appndx_image_1nn}.

\section{Experiments on text representations}
\begin{table*}[!t]
    \centering
    \renewcommand{\arraystretch}{1.3}
    \resizebox{\textwidth}{!}{
    \begin{tabular}{c|ccccccc|c}
    \toprule
    \toprule
         \textbf{Dim} & \textbf{Full Rep.} & \textbf{Rand. LP} & \textbf{SVD} & \textbf{MRL} & \textbf{MRL-E} & \textbf{Unsup. HERL} & \textbf{Sup. HERL} & \textbf{CSR}\\
         \midrule
        8 & - & 4.56 & 2.34 & 66.63 & 56.66 & 73.52 & \textbf{74.16} & 73.62 \\
        16 & - & 11.29 & 7.17 & 73.53 & 71.94 & 74.75 & \textbf{75.04} & 74.75 \\
        32 & - & 27.21 & 20.46 & 75.03 & 74.48 & 75.48 & \textbf{76.02 }& 75.44 \\
        64 & - & 49.47 & 48.10 & 75.82 & 75.35 & 76.01 & \textbf{76.53} & 75.88\\
        128 & - & 65.70 & 67.24 & 76.30 & 75.80 & 76.29 & \textbf{76.88} & 76.24 \\
        256 & - & 72.43 & 74.59 & 76.47 & 76.22 & 76.55 & \textbf{77.01} & 76.25\\
        512 & - & 74.94 & 76.78 & 76.65 & 76.36 & 76.78 & \textbf{77.18} & 76.34 \\
        1024 & - & 76.10 & 76.87 & 76.76 & 76.48 & 76.99 & \textbf{77.31} & 76.54 \\
        2048 & 76.87 & 76.87 & - & 76.80 & 76.51 & 77.03 & \textbf{77.45} & 76.52 \\
        \bottomrule
        \bottomrule
    \end{tabular}}
    \caption{\textbf{Top-1 accuracy.} Comparison of different methods on ImageNet-1k using a linear probe for classification. Note that, CSR \cite{csr} uses an embedding dimension of 8192 (4x) and activate only certain dimensions. Hence, has been omitted from comparison. The highest score has been \textbf{bolded}.}
    \label{tab:appndx_image_cls}
\end{table*}
\subsection{Datasets}
\label{subsec:text_datasets}
We evaluate on a broad slice of the Massive Text Embedding
Benchmark~\cite{muennighoff2022mteb}, covering three task families
that stress different aspects of an elastic embedding: classification,
clustering, and retrieval. All datasets are accessed through the MTEB
release; for brevity we cite their original sources below and refer
the reader to \cite{muennighoff2022mteb} for the exact splits and
evaluation protocols.
 
\paragraph{Classification.}
We report accuracy on three datasets spanning intent detection,
sentiment, and topic prediction in both English and multilingual
settings.
\textsc{MTOPIntent}~\cite{li2021mtop} is a multilingual
task-oriented parsing benchmark with $\sim$100{,}000 utterances over
six languages and eleven domains.
\textsc{Banking77}~\cite{casanueva2020banking77} contains
13{,}083 customer-service queries labelled with 77 fine-grained
banking intents.
\textsc{TweetSentimentExtraction}~\cite{maggie2020tweet} provides
ternary sentiment labels (positive / neutral / negative) on tweets,
released originally as a Kaggle competition.
 
\paragraph{Clustering.}
We further evaluate on three clustering tasks at varying granularity
and document length.
 \textsc{BiorxivClusteringS2S}, and
\textsc{BiorxivClusteringP2P}~\cite{muennighoff2022mteb} cluster
papers by category using either titles alone (S2S) or
title-plus-abstract (P2P), exposing the adaptor to short and long
inputs respectively.
\textsc{TwentyNewsgroupsClustering}~\cite{lang1995newsweeder}
clusters 20-class newsgroup titles across ten splits of 1{,}000 to
10{,}000 documents.
 
\paragraph{Retrieval.}
We measure nDCG@10 on three retrieval tasks covering domain QA,
fact-checking, argumentation, and paraphrase identification.
\textsc{FiQA-2018}~\cite{maia2018fiqa} is an aspect-based
sentiment and opinion QA benchmark in the financial domain.
\textsc{NFCorpus}~\cite{boteva2016nfcorpus} contains 3{,}244
natural-language medical queries with 169{,}756 relevance judgments
over 9{,}964 documents.
\textsc{SciFact}~\cite{wadden2020scifact} pairs 1{,}400
expert-written scientific claims with evidence-containing abstracts
annotated for veracity and rationales.

\subsection{Implementation Details}
\label{subsec:text_implement}

We follow CSR \cite{csr} and utilize NV-Embed-V2 \cite{lee2025nvembedimprovedtechniquestraining} to extract encoder representations for our HERL model, adopting the same preprocessing procedures as well. All experiments have been performed on a single H100 GPU. Note that, we also clip the norm of the representations to $5.0$ before Lorentzian projection to avoid numerical instability errors. For larger dimensions, we perform graident accumulation to allow operating at the batch size. All hyperbolic operations have been performed using the $\mathrm{geoopt}$ library. See Table \ref{tab:text_impl_details} for more specific details. For classification tasks, the projector module is an MLP made of four layers ($4096 \to 2048 \to 1024 \to 512$). Clustering tasks use a smaller network ($4096 \to 2048 \to 1024$). On the other hand, retrieval tasks utilize the same architecture as classification tasks.

\begin{table*}[!b]
    \centering
    \renewcommand{\arraystretch}{1.3}
    \resizebox{\textwidth}{!}{
    \begin{tabular}{c|ccccccc}
    \toprule
    \toprule
    Dataset & Backbone & Input dim & LR & Epochs & Batch size & Optimizer & Curvature \\
    \midrule
   MTOPIntent & \multirow{3}{*}{NV-Embed-V2} & \multirow{3}{*}{4096} & \multirow{3}{*}{4e-5} & \multirow{3}{*}{10} & \multirow{3}{*}{128} & \multirow{3}{*}{RiemannianAdam} & \multirow{3}{*}{0.9} \\
   Banking77 &  &  &  &  &  & & \\
   TweetSentiment &  &  &  &  &  &  & \\
   \midrule
    BiorxivP2P& \multirow{3}{*}{NV-Embed-V2} & \multirow{3}{*}{4096} & 1e-3 & \multirow{3}{*}{10} & \multirow{3}{*}{128} & \multirow{3}{*}{RiemannianAdam} & \multirow{3}{*}{0.9} \\
    
     BiorxivS2S&  & & 1e-3 &  &  &  &  \\
      TwentyNews&  &  & 1e-3 &  &  &  &  \\
      \midrule
       FIQA & \multirow{3}{*}{NV-Embed-V2} &  \multirow{3}{*}{4096} & 1e-3 & \multirow{3}{*}{10} & \multirow{3}{*}{256} & \multirow{3}{*}{RiemannianAdam} & 1.5 \\
        NFCorpus &  &  & 1e-3 &  &  &  & 0.8 \\
         SciFACT&  &  & 1e-3 &  &  & & 1.0 \\
    \bottomrule
    \bottomrule
    \end{tabular}}
    \caption{Implementation details for experiments on text representations across nine datasets.}
    \label{tab:text_impl_details}
\end{table*}

\subsection{Results}
Exact numerical values for all tasks: classification, clustering and retrieval are shown in Table \ref{tab:appndx_text_results}.
\label{subsec: text_results_appndx}
\begin{table*}[t]
\centering
\renewcommand{\arraystretch}{1.15}
\resizebox{\textwidth}{!}{
\begin{tabular}{llcccc|ccc|ccc}
\toprule
\toprule
& & \makecell{Dim} & \multicolumn{3}{c|}{\makecell{\textbf{Text Classification}\\Top-1 Acc (\%)$\uparrow$}}
& \multicolumn{3}{c|}{\makecell{\textbf{Text Clustering}\\Top-1 Acc (\%)$\uparrow$}}
& \multicolumn{3}{c}{\makecell{\textbf{Text Retrieval}\\NDCG@10 (\%)$\uparrow$}} \\
\cmidrule(lr){4-6} \cmidrule(lr){7-9} \cmidrule(lr){10-12}
\textbf{Category} & \textbf{Model} & &
MTOPIntent & Banking77 & TweetSentiment &
BiorxivP2P & BiorxivS2S & TwentyNews &
FiQA2018 & NFCorpus & SciFACT \\
\midrule

Full Rep & NV-Embed-V2 & 4096 
& 93.58 & 92.20 & 79.73
& 53.61 & 49.60 & 64.82
& 62.65 & 43.97 & 77.93 \\
\midrule

\multirow{9}{*}{MRL}
& Stella-1.5B-v5 & 256 
& 90.45 & 86.14 & 76.75
& 50.81 & 46.42 & 60.07
& 55.59 & 36.97 & \textbf{77.48} \\
& Jina-v3 & 256 
& 78.81 & 84.08 & 73.81
& 38.14 & 34.39 & 51.96
& 55.73 & 36.63 & 66.63 \\
& Nomic-Embed-V1.5 & 256 
& 72.47 & 83.69 & 59.20
& 38.19 & 31.83 & 48.56
& 35.00 & 32.54 & 68.24 \\
& Gecko-Embed-004(Google) & 256 
& 77.82 & 86.01 & 72.97
& 36.28 & 33.09 & 50.60
& 55.54 & 37.81 & 70.86 \\
& Text-Embed-3-L (OpenAI) & 256 
& 70.45 & 83.19 & 58.98
& 35.43 & 33.86 & 54.24
& 50.33 & 37.94 & 73.10 \\
& Arctic-Embed-L-V2 & 256 
& 67.69 & 80.99 & 59.06
& 34.25 & 34.07 & 30.06
& 44.69 & 35.02 & 69.51 \\
& M2V-Base-Glove & 256 
& 59.26 & 72.39 & 50.02
& 32.26 & 22.34 & 25.38
& 11.82 & 23.15 & 50.66 \\
& Jina-V3 & 64 
& 68.12 & 67.98 & 71.18
& 36.89 & 33.57 & 50.22
& 44.18 & 33.66 & 68.84 \\
& Nomic-Embed-V1.5 & 64 
& 62.77 & 80.63 & 55.23
& 34.81 & 44.61 & 48.06
& 10.22 & 18.96 & 36.55 \\
& Potion-Base-2M & 64 
& 42.50 & 65.17 & 52.52
& 25.78 & 14.94 & 27.07
& 32.08 & 30.72 & 64.28 \\
\midrule

\rowcolor{gray!15}
Ours & Unsup. HERL & 32 
& 94.39 & 92.88 & 79.51
& 48.18 & 48.47 & 62.85
& 50.70 & 45.67 & 55.70 \\
\rowcolor{gray!15}
(w/ NV-Embed-V2) & Unsup. HERL & 8192 
& 93.72 & 93.21 & 79.34
& 78.43 & 75.22 & 64.70
& \textbf{61.11} & \textbf{55.23} & 71.62 \\
\rowcolor{gray!15}
 & Sup. HERL & 32 
& \textbf{94.53} & \textbf{94.86} & \textbf{79.88}
& 63.28 & 75.82 & 80.14
& - & - & - \\
\rowcolor{gray!15}
 & Sup. HERL & 8192
& 93.49 & 93.15 & 79.60
& \textbf{80.12} & \textbf{76.29} & \textbf{82.77}
& - & - & - \\
\midrule
Sparse & CSR & 32 
& 89.86 & 91.02 & 78.55
& 53.49 & 49.13 & 63.05
& 57.54 & 38.06 & 71.17 \\
\bottomrule
\bottomrule
\end{tabular}}
\caption{\textbf{Results on textual representations.} We compare the downstream utility of representations across three tasks: classification, clustering and retrieval. Note that, CSR \cite{csr} uses a 4x ambient dimension, activating only certain dimensions. Hence, has been omitted from comparison. The highest score has been \textbf{bolded}.}
    \label{tab:appndx_text_results}
\end{table*}

\section{Effect of Projector Module}
We also explored a tied encoder–decoder architecture without the projector module. Empirically, this configuration led to a substantial degradation in performance for both low-dimensional representations and expanded high-dimensional representations beyond the encoder dimensionality, even under extensive fine-tuning. This suggests that a single linear weight matrix lacks the expressive capacity required to effectively map Euclidean representations onto hyperbolic manifolds, thereby limiting the model’s ability to fully exploit the geometric structure and representational advantages of hyperbolic space.

\newpage
\input{checklist.tex}

\end{document}

%% file: checklist.tex
\section*{NeurIPS Paper Checklist}

The checklist is designed to encourage best practices for responsible machine learning research, addressing issues of reproducibility, transparency, research ethics, and societal impact. Do not remove the checklist: {\bf The papers not including the checklist will be desk rejected.} The checklist should follow the references and follow the (optional) supplemental material.  The checklist does NOT count towards the page
limit. 

Please read the checklist guidelines carefully for information on how to answer these questions. For each question in the checklist:
\begin{itemize}
    \item You should answer \answerYes{}, \answerNo{}, or \answerNA{}.
    \item \answerNA{} means either that the question is Not Applicable for that particular paper or the relevant information is Not Available.
    \item Please provide a short (1--2 sentence) justification right after your answer (even for \answerNA). 
\end{itemize}

{\bf The checklist answers are an integral part of your paper submission.} They are visible to the reviewers, area chairs, senior area chairs, and ethics reviewers. You will also be asked to include it (after eventual revisions) with the final version of your paper, and its final version will be published with the paper.

The reviewers of your paper will be asked to use the checklist as one of the factors in their evaluation. While \answerYes{} is generally preferable to \answerNo{}, it is perfectly acceptable to answer \answerNo{} provided a proper justification is given (e.g., error bars are not reported because it would be too computationally expensive'' or ``we were unable to find the license for the dataset we used''). In general, answering \answerNo{} or \answerNA{} is not grounds for rejection. While the questions are phrased in a binary way, we acknowledge that the true answer is often more nuanced, so please just use your best judgment and write a justification to elaborate. All supporting evidence can appear either in the main paper or the supplemental material, provided in appendix. If you answer \answerYes{} to a question, in the justification please point to the section(s) where related material for the question can be found.

IMPORTANT, please:
\begin{itemize}
    \item {\bf Delete this instruction block, but keep the section heading ``NeurIPS Paper Checklist"},
    \item  {\bf Keep the checklist subsection headings, questions/answers and guidelines below.}
    \item {\bf Do not modify the questions and only use the provided macros for your answers}.
\end{itemize}


\begin{enumerate}

\item {\bf Claims}
    \item[] Question: Do the main claims made in the abstract and introduction accurately reflect the paper's contributions and scope?
    \item[] Answer: \answerYes{} 
    \item[] Justification: We propose a novel framework, Hyperbolic Elastic Representation Learning (HERL) to learn nested and elastic representations without retraining the encoder, unlike Matryoshka Representation Learning that requires encoder retraining.

\item {\bf Limitations}
    \item[] Question: Does the paper discuss the limitations of the work performed by the authors?
    \item[] Answer: \answerYes{} 
    \item[] Justification: HERL requires aggressive finetuning on retrieval tasks due to the metric mismatch. To keep consistent with evaluation metrics that normalize representations (project to hypersphere) before computing 1-NN accuracy, we report scores using the same protocol. However, normalizing hyperbolic optimized representations erase radial information, which is a primal source of separation in hyperbolic geometry. This has been substantiated in the main paper.

\item {\bf Theory assumptions and proofs}
    \item[] Question: For each theoretical result, does the paper provide the full set of assumptions and a complete (and correct) proof?
    \item[] Answer: \answerYes{} 
    \item[] Justification: Theory backed by emperical results has been shown in the main paper and appendix.

    \item {\bf Experimental result reproducibility}
    \item[] Question: Does the paper fully disclose all the information needed to reproduce the main experimental results of the paper to the extent that it affects the main claims and/or conclusions of the paper (regardless of whether the code and data are provided or not)?
    \item[] Answer: \answerYes{} 
    \item[] Justification: Yes, we provide dataset-specific details to allow full reproducibility. Further, code will be released upon acceptance.

\item {\bf Open access to data and code}
    \item[] Question: Does the paper provide open access to the data and code, with sufficient instructions to faithfully reproduce the main experimental results, as described in supplemental material?
    \item[] Answer: \answerYes{} 
    \item[] Justification: We do not propose or create a new dataset. All datasets are publicly available and code will be released upon acceptance.

\item {\bf Experimental setting/details}
    \item[] Question: Does the paper specify all the training and test details (e.g., data splits, hyperparameters, how they were chosen, type of optimizer) necessary to understand the results?
    \item[] Answer: \answerYes{} 
    \item[] Justification: Yes, dataset-specific hyperparameters have been discussed in appendix.

\item {\bf Experiment statistical significance}
    \item[] Question: Does the paper report error bars suitably and correctly defined or other appropriate information about the statistical significance of the experiments?
    \item[] Answer: \answerYes{} 
    \item[] Justification: We have carefully followed the protocols from prior works in reporting all metrics.

\item {\bf Experiments compute resources}
    \item[] Question: For each experiment, does the paper provide sufficient information on the computer resources (type of compute workers, memory, time of execution) needed to reproduce the experiments?
    \item[] Answer: \answerYes{} 
    \item[] Justification: All experiments have been completed on a single H100 GPU, this has been mentioned in the appendix as well.
    
\item {\bf Code of ethics}
    \item[] Question: Does the research conducted in the paper conform, in every respect, with the NeurIPS Code of Ethics \url{https://neurips.cc/public/EthicsGuidelines}?
    \item[] Answer: \answerYes{} 
    \item[] Justification: Yes, we adhere to the NeurIPS Code of Ethics and preserve anonymity throughout the paper.

\item {\bf Broader impacts}
    \item[] Question: Does the paper discuss both potential positive societal impacts and negative societal impacts of the work performed?
    \item[] Answer: \answerYes{} 
    \item[] Justification: We have shown throughout the paper that HERL allows efficient and effective representations, saving hours of training time. Thereby, drastically reducing energy requirements.
    
\item {\bf Safeguards}
    \item[] Question: Does the paper describe safeguards that have been put in place for responsible release of data or models that have a high risk for misuse (e.g., pre-trained language models, image generators, or scraped datasets)?
    \item[] Answer: \answerNA{} 
    \item[] Justification: N/A

\item {\bf Licenses for existing assets}
    \item[] Question: Are the creators or original owners of assets (e.g., code, data, models), used in the paper, properly credited and are the license and terms of use explicitly mentioned and properly respected?
    \item[] Answer: \answerYes{} 
    \item[] Justification: Yes, we have cited all dataset sources, code and models.

\item {\bf New assets}
    \item[] Question: Are new assets introduced in the paper well documented and is the documentation provided alongside the assets?
    \item[] Answer: \answerNA{} 
    \item[] Justification: We do not release asset along with the paper.

\item {\bf Crowdsourcing and research with human subjects}
    \item[] Question: For crowdsourcing experiments and research with human subjects, does the paper include the full text of instructions given to participants and screenshots, if applicable, as well as details about compensation (if any)? 
    \item[] Answer: \answerNA{} 
    \item[] Justification: No experiments involved crowdsourcing or human subjects.

\item {\bf Institutional review board (IRB) approvals or equivalent for research with human subjects}
    \item[] Question: Does the paper describe potential risks incurred by study participants, whether such risks were disclosed to the subjects, and whether Institutional Review Board (IRB) approvals (or an equivalent approval/review based on the requirements of your country or institution) were obtained?
    \item[] Answer: \answerNA{} 
    \item[] Justification: N/A

\item {\bf Declaration of LLM usage}
    \item[] Question: Does the paper describe the usage of LLMs if it is an important, original, or non-standard component of the core methods in this research? Note that if the LLM is used only for writing, editing, or formatting purposes and does \emph{not} impact the core methodology, scientific rigor, or originality of the research, declaration is not required.
    \item[] Answer: \answerNA{} 
    \item[] Justification: N/A

\end{enumerate}